\documentclass[10pt,twocolumn,letterpaper]{article}

\usepackage{iccv}
\usepackage{footnote}
\newcommand\blfootnote[1]{
  \begingroup
  \renewcommand\thefootnote{}\footnote{#1}
  \addtocounter{footnote}{-1}
  \endgroup
}
\usepackage{graphicx}
\usepackage{amsmath}
\usepackage{amssymb}
\usepackage{booktabs}
\usepackage{times}
\usepackage{epsfig}
\usepackage{graphicx}
\usepackage{amsmath}
\usepackage{amssymb}
\usepackage{array}
\usepackage{csquotes}
\usepackage[accsupp]{axessibility} %

\usepackage[numbers,sort]{natbib}
\usepackage{xcolor, colortbl}
\usepackage{tabularx}
\usepackage{multirow}
\usepackage{enumitem}
\usepackage{bbm}
\usepackage{pifont}
\usepackage{capt-of}
\usepackage{array,multirow}
\usepackage{booktabs,siunitx,caption}
\usepackage[export]{adjustbox}
\usepackage{lipsum} 
\usepackage{duckuments}

\newcolumntype{L}[1]{>{\raggedright\let\newline\\\arraybackslash\hspace{0pt}}m{#1}}
\newcolumntype{C}[1]{>{\centering\let\newline\\\arraybackslash\hspace{0pt}}m{#1}}
\newcolumntype{R}[1]{>{\raggedleft\let\newline\\\arraybackslash\hspace{0pt}}m{#1}}

\def\ie{\emph{i.e.}}
\def\eg{\emph{e.g.}}

\definecolor{Gray}{gray}{0.85}
\definecolor{brown}{rgb}{0.65, 0.16, 0.16}
\definecolor{purple}{rgb}{0.65, 0.1, 0.75}
\definecolor{yellow}{rgb}{0.95, 0.9, 0.0}
\definecolor{blond}{rgb}{0.98, 0.94, 0.75}
\definecolor{orange}{rgb}{1.0, 0.5, 0.2}
\definecolor{dorange}{rgb}{0.9, 0.3, 0.3}
\definecolor{green}{rgb}{0, 0.7, 0.4}
\definecolor{red}{rgb}{0.8, 0.0, 0.0}
\definecolor{dblue}{rgb}{0.6, 0.1, 0.9}
\definecolor{grey}{rgb}{0.9, 0.9, 0.9}

\newcommand{\Fig}[1]{Fig.~\ref{fig:#1}}
\newcommand{\Sec}[1]{Sec.~\ref{sec:#1}}
\newcommand{\Eq}[1]{Eq.~(\ref{eq:#1})}
\newcommand{\Tbl}[1]{Table~\ref{tab:#1}}

\newcommand{\cmark}{\ding{51}}%
\newcommand{\xmark}{\ding{55}}%

\newcommand{\ccol}{\cellcolor{grey}}

\usepackage[pagebackref=true,breaklinks=true,letterpaper=true,colorlinks,bookmarks=false]{hyperref}

\usepackage[capitalize]{cleveref}
\crefname{section}{Sec.}{Secs.}
\Crefname{section}{Section}{Sections}
\Crefname{table}{Table}{Tables}
\crefname{table}{Tab.}{Tabs.}

\iccvfinalcopy %

\def\iccvPaperID{3753} %

\ificcvfinal\fi

\usepackage{xcolor, colortbl}
\usepackage{subcaption,booktabs}

\begin{document}

\title{Shatter and Gather:\\Learning Referring Image Segmentation with Text Supervision \vspace{-1mm}}

\author{
Dongwon Kim$^{1\ast}$ \hspace{4mm} Namyup Kim$^{1\ast}$ \hspace{4.5mm} Cuiling Lan$^2$ \hspace{4.5mm} Suha Kwak$^1$\\
$^1$POSTECH \hspace{8mm} $^2$Microsoft Research Asia \\
{\small $^1$\tt \{kdwon, namyup, suha.kwak\}@postech.ac.kr \hspace{4.5mm} $^2$culan@micrsoft.com} \\
}

\maketitle
\ificcvfinal\thispagestyle{empty}\fi

\begin{abstract}
Referring image segmentation, the task of segmenting any arbitrary entities described in free-form texts, opens up a variety of vision applications.
However, manual labeling of training data for this task is prohibitively costly, leading to lack of labeled data for training.
We address this issue by a weakly supervised learning approach using text descriptions of training images as the only source of supervision.
To this end, we first present a new model that discovers semantic entities in input image and then combines such entities relevant to text query to predict the mask of the referent.
We also present a new loss function that allows the model to be trained without any further supervision. 
Our method was evaluated on four public benchmarks for referring image segmentation, where it clearly outperformed the existing method for the same task and recent open-vocabulary segmentation models on all the benchmarks.
\end{abstract}
\vspace{-5mm}

\blfootnote{$^{\ast}$Equal contribution.}

\vspace{-1mm}
\section{Introduction}
\vspace{-1.5mm}
\label{sec:intro}

Referring image segmentation is the task of segmenting the referent corresponding to a natural language expression given as a query.
Unlike the conventional semantic segmentation that aims at segmenting a pre-defined set of classes, referring image segmentation enables segmentation of any arbitrary entities described in free-form texts and thus opens up a wide variety of applications such as human-computer interaction~\cite{wang2019reinforced, chen2021history} and text-based image editing~\cite{chen2018language, fu2022m3l}.

Thanks to the development of deep neural networks, recent studies have demonstrated remarkable results on referring image segmentation in the supervised learning setting~\cite{hu2016segmentation, yu2018mattnet, ye2019cross, hui2020linguistic, huang2020referring, ding2021vision, yang2022lavt}.
However, {obtaining} manual annotation of training data for referring image segmentation is prohibitively costly since the data requires two types of labels for each image, {\ie,} natural language expressions describing the entities that appear in the image and segmentation labels corresponding to them. 
{This} typically demands tremendously more manpower than annotation for semantic segmentation,
leading to lack of labeled data for training.

\begin{figure} [!t]
\centering
\includegraphics[width=\columnwidth]{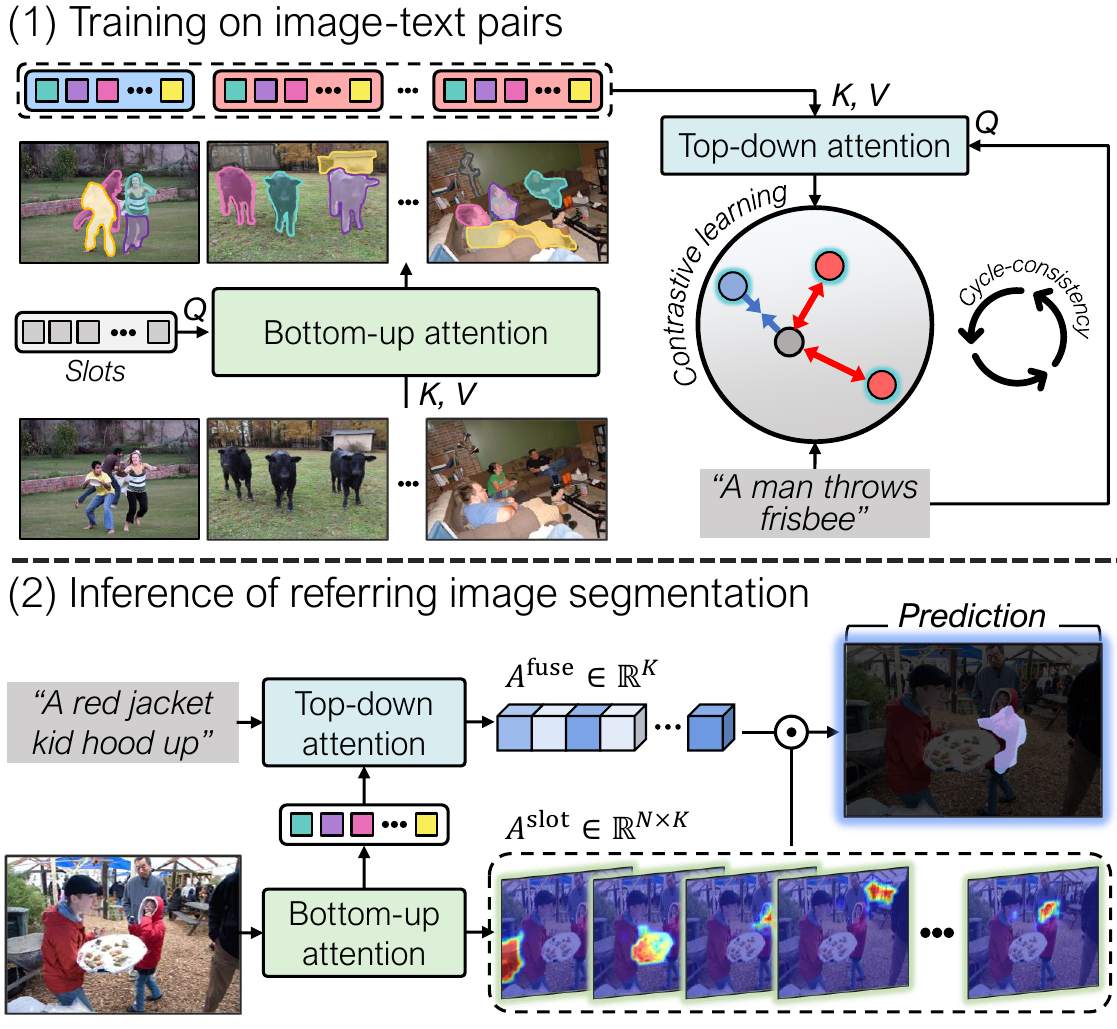}
\vspace{-6mm}
\caption{
Our model consists of two different attention modules: bottom-up attention that identifies individual entities in an image, and top-down attention that combines the entities based on a referring expression.
It is trained in a contrastive manner to ensure consistency between matching image-text pairs. 
In inference, a segmentation mask is predicted by combining the entities found by bottom-up attention with weights derived by top-down attention.
}
\vspace{-3mm}
\label{fig:teaser}
\end{figure}

A solution to this issue is weakly supervised learning, which trains a model on a dataset providing weaker forms of supervision than the conventional ones.
In this paper, we particularly focus on learning referring image segmentation only with natural language expressions describing {the} associated images.
Since these labels are readily available in many vision-language datasets~\cite{radford2021learning, jia2021scaling}, the issue on lack of labeled training data can be alleviated by our approach.
However, learning referring image segmentation in this setting introduces another challenge:
Since natural language expressions often contain relations between visual entities such as instances and object parts (\eg, ``The cat next to the table leg''), a model to be trained should be aware of the inter-object relations to exploit the expressions as supervision.
Thereby we need a method to discover individual entities and infer their relations without any supervision but the referring expression.

To fulfill this requirement, we introduce a new \textit{bottom-up} and \textit{top-down} attention framework, which is illustrated in \Fig{teaser}.
Individual entities existing in an image are first discovered by \textit{bottom-up} attention, which solely exploits visual information.
For the {bottom-up} attention, an entity discovery module progressively refines a set of embedding vectors named \emph{slots} to capture distinct visual entities, following slot attention~\cite{locatello2020object}.
We moreover propose a novel slot formulation named \textit{entity slot}, which enables fine-grained entity discovery 
in real-world images.
As a result, entity slots after the refinement capture individual visual entities without any segmentation supervision. These entities are considered primitive units for 
composing the segmentation mask of a referent during inference.

Afterward, based on the relational and structural information of referring expression, {top-down} attention selectively attends to the referred entities and combines them into the predicted mask.
{Top-down} attention is implemented by a modality fusion module consists of cross-attention transformer~\cite{dosovitskiy2020image}.
This attention scheme enables the module to infer the relevance between discovered entities and the referring expression instead of relying on the cosine similarity or heuristic similarity functions used in previous work~\cite{strudel2022weakly,ghiasi2022scaling,xu2022groupvit,zhou2022extract}.

For training, we propose the contrastive cycle-consistency (C$^{3}$) loss that enforces cycle-consistency between image-text pairs under the contrastive learning framework~\cite{radford2021learning}.
To establish such consistency, the model is trained to preserve the textual information after the fusion of entities and textual features.
By doing so, we observe that latent relevance between the discovered entities and referring expression automatically emerges in the {top-down} attention without any explicit mask supervision. 

One may wonder the difference between our task and open-vocabulary segmentation~\cite{xu2022groupvit, ghiasi2022scaling, zhou2022extract}, \ie, semantic segmentation of arbitrary categories without explicit mask supervision.
The key difference is that a referent to segment in referring image segmentation is given by a complex free-form text, which introduces additional challenges.

The proposed method was evaluated and compared on four public benchmarks for referring image segmentation, where it substantially outperformed the previous weakly supervised learning method~\cite{strudel2022weakly}.
Moreover, we reproduced recent open-vocabulary segmentation models~\cite{xu2022groupvit, zhou2022extract} and evaluated them on referring image segmentation benchmarks;
the results show that the proposed model surpassed them even without pre-training on large-scale image-text data~\cite{changpinyo2021conceptual, thomee2016yfcc100m, radford2021learning}.

In summary, our contribution is three-fold as follows:
\begin{itemize}%
   \setlength\itemsep{-0.5mm}
   \vspace{-2mm}\item {We present a new weakly supervised learning model for referring image segmentation. Our model employs bottom-up and top-down attention to discover individual entities and infer their relations, which enables to exploit natural language expressions as supervision effectively during training.}
   \vspace{-0.5mm}\item {We propose a new loss function called contrastive cycle-consistency loss, which allows latent {relevance} between discovered entities and referring expressions to emerge without requiring further supervision.}
   \vspace{-0.5mm}\item {Our model clearly outperformed the existing method for the same task and recent open-vocabulary segmentation models on all the four benchmarks.}
\end{itemize}
\vspace{-1mm}

\vspace{-1mm}
\section{Related Work}
\vspace{-1.5mm}
\label{sec:relatedwork}

\begin{figure*}[t!]
    \centering
    \includegraphics[width=\linewidth]{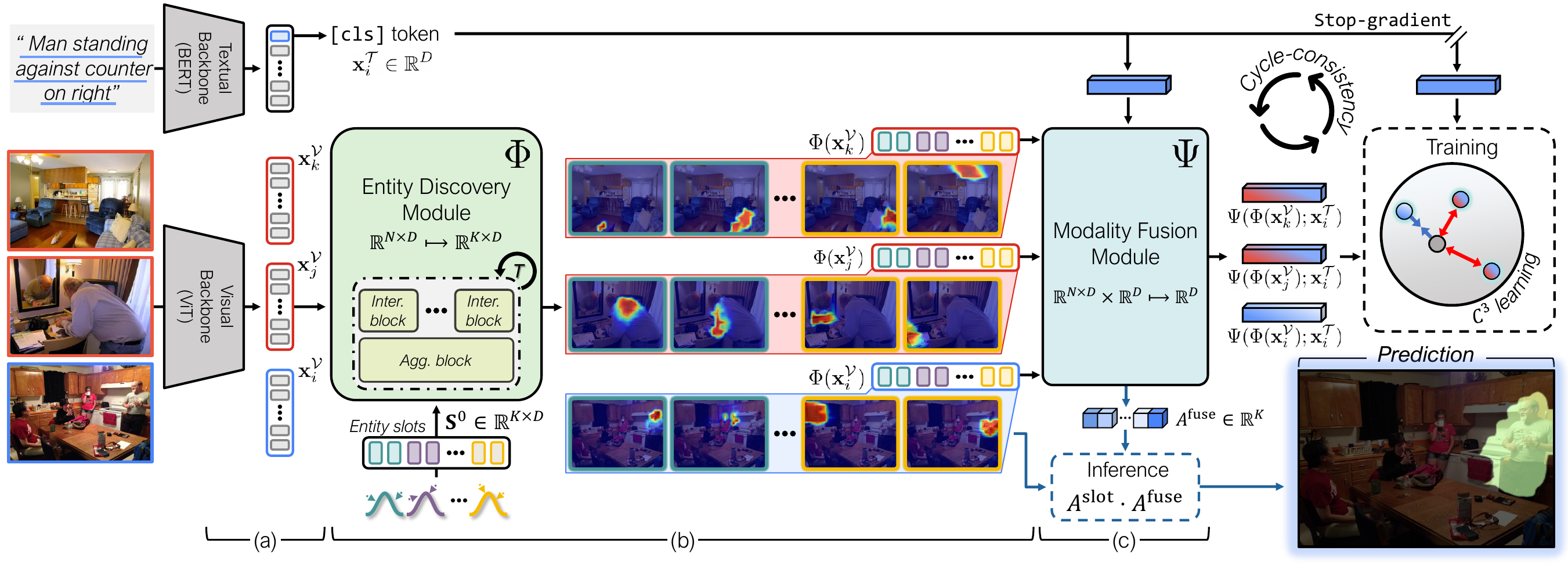}
\vspace{-5.5mm}
\caption{
Illustration of the overall architecture of our model along with its behavior during training and inference.
(a) Feature extractors: The visual and textual features are extracted by transformer-based encoders of the two modalities respectively.
(b) Entity discovery module: A set of visual entities are discovered from the visual features through the bottom-up attention.
(c) Modality fusion module: The visual entities and a referring expression are fused considering their relevance, which is estimated by the top-down attention.
Our model is trained with the proposed contrastive cycle-consistency (C$^3$) loss between the cross-modal embeddings and textual features. In inference, the segmentation mask is predicted by jointly considering the visual entities $A^{\text{slot}}$ from the entity discovery module and the relevance scores $A^{\text{fuse}}$ from the modality fusion module.
} \label{fig:overall}
\vspace{-3mm}
\end{figure*}

\noindent \textbf{Referring image segmentation:}
Early approaches~\cite{hu2016segmentation, liu2017recurrent, li2018referring, shi2018key} obtain multimodal feature maps by simple concatenation-convolution operation.
With this operation as a basis, most existing studies have suggested attention mechanisms and multi-level feature aggreations~\cite{margffoy2018dynamic, chen2019see, ye2019cross, hu2020bi, feng2021encoder, huang2020referring}.
Various approaches have been employed for calculating fine interactions between the modalities, including multimodal feature fusion~\cite{kamath2021mdetr, ding2021vision, kim2022restr} using transformers~\cite{dosovitskiy2020image}, a large-scale vision-language pretrained framework~\cite{wang2022cris}, and holistically multimodal fusion~\cite{yang2022lavt}.
The existing referring image segmentation methods have been developed in a fully supervised learning setting.
Recently, a weakly supervised learning method~\cite{strudel2022weakly} using only image-text pairs for training is introduced to alleviate the high annotation cost.
However, this study relies on a heuristic similarity function to infer relations between patches and texts, which suffers from comprehending the relational and structural information from image and text modalities.
Therefore, we propose a new weakly supervised model consisting of bottom-up and top-down attention to discover visual entities and infer their relations effectively.

\vspace{0.5mm} \noindent \textbf{Phrase grounding:}
Phrase grounding~\cite{rohrbach2016grounding, yu2018mattnet, huang2021look, kamath2021mdetr, deng2021transvg, yang2022improving} is the task of learning the association between image regions and words in caption describing the image.
Weakly supervised approaches have also been proposed for this task~\cite{xiao2017weakly, datta2019align2ground, gupta2020contrastive, wang2021improving, liu2021relation}, where the correspondence between words and regions are not given.
It is worth noting that the contrastive learning framework introduced in~\cite{gupta2020contrastive} is closely related to the proposed C$^3$ loss by enforcing mutual information between regions and words. 
However, these studies rely on an off-the-shelf object detector~\cite{Rcnn}, where the final prediction is restricted by the predefined set of classes with which the detector is trained.
On the contrary, our method discovers visual entities in an image without an auxiliary model trained with external labeled data, which is more suitable for referring image segmentation.

\vspace{0.5mm} \noindent \textbf{Weakly supervised learning for segmentation:}
Weakly supervised learning has become a favored solution for addressing the lack of annotated data in training, especially for semantic segmentation~\cite{sec, Huang_2018_CVPR, Hong2017_webly, sppnet, IRNet, affinitynet, zhang2020causal, xu2022multi, zou2021pseudoseg} and instance segmentation~\cite{PRM, SDI, Cutnpaste, kim2022beyond}.
These methods refine coarse localization cues obtained by CAMs~\citep{Cam,Selvaraju_2017_ICCV,Wei_2017_CVPR}, and use the results as pseudo labels for self-training.
However, they cannot be directly applied to the referring image segmentation due to the following challenges.
First, there are no predefined classes in this task, making it difficult to learn shared semantics within CAMs by multiple instance learning.
Second, there is no mechanism to discover visual entities and combine them with textual information.

\vspace{0.5mm} \noindent \textbf{Slot attention:} 
Slot attention~\cite{locatello2020object} is proposed for the learning object-centric representation.
It can discover the objects within an image in an unsupervised manner and be used in the various downstream task such as segmentation~\cite{xu2022groupvit,zhou2022slot} and retrieval~\cite{kim2022improving, Weinzaepfel2022LearningSF}.
In specific, slot attention introduces an iterative attention mechanism that operates on the slots, which are the set of embedding vectors corresponding to the objects that appear in the input data.
In previous work, slots are initialized as a set of learnable embeddings~\cite{xu2022groupvit,zhou2022slot,kim2022improving,Weinzaepfel2022LearningSF} or a set of vectors sampled from a learnable Gaussian distribution~\cite{locatello2020object,singh2021illiterate}.
However, we empirically observed that both types of initialization are inapplicable to visual entity discovery for referring image segmentation, which we will elaborate on in \Sec{slotattn}.

\vspace{0.5mm} \noindent \textbf{Open-vocabulary segmentation:} 
The open-vocabulary segmentation is proposed to address the limitation of semantic segmentation that it is constrained to the pre-defined set of classes.
Unlike semantic segmentation, open-vocabulary segmentation models are evaluated with classes unseen during training.
The early approaches propose a generative model~\cite{bucher2019zero} and a projection model~\cite{xian2019semantic} to generate or project pixel-level semantic features from visual features for the unseen classes through word embedding space.
Recent studies~\cite{xu2022groupvit, zhou2022extract, li2022languagedriven, ghiasi2022scaling} largely improve the quality of semantic features for the unseen classes, and further introduce the methods that do not require segmentation labels for seen classes using large-scale image-text pretrained models (\eg CLIP~\cite{radford2021learning}, ALIGN~\cite{jia2021scaling}).
However, since they are trained to predict masks for semantic categories, not for free-form texts describing arbitrary entities in an image, their applicability is still limited.
We thus propose a new method for referring image segmentation that only requires image-text pairs during training.

\vspace{-1mm}
\section{Proposed Method}
\vspace{-1.5mm}

As shown in Fig.~\ref{fig:overall}, our model consists of three main components: feature extractors, entity discovery module, and modality fusion module. 
The visual and textual features are extracted from the input using the backbone networks of the two modalities respectively. (\Sec{feature}).
{The visual features are then given as input to the entity discovery module, which aggregates the features into multiple entity embeddings using the bottom-up attention (\Sec{slotattn}).
The entity embeddings correspond to distinct entities appearing in the image, which are considered as the smallest units of building the segmentation mask.}
{The modality fusion module fuses the entity embeddings and the textual feature into cross-modal embeddings according to their relevance using the top-down attention (\Sec{crossattn}).}
The produced cross-modal embedding is used only for training, with a learning objective enforcing cycle-consistency and reconstruction of visual features (\Sec{train}).
{The segmentation mask is predicted by combining the entities discovered by the entity discovery module with the relevance scores estimated by the modality fusion module (\Sec{infer}).}

\subsection{Visual and Textual Feature Extraction}
\vspace{-1mm}
\label{sec:feature}
Here, we provide the details of the transformer-based backbone network for each modality.

\vspace{0.5mm} \noindent \textbf{Visual feature extraction:}
We employ the vision transformer (ViT)~\cite{dosovitskiy2020image} as the visual backbone network.
An input image is first split into $N$ patches and then linearly projected.
The patches are processed by the multiple self-attention blocks of ViT, where we refer to the output tokens of the last block as the visual features $\mathbf{x}^{\mathcal{V}} \in \mathbb{R}^{N \times D}$.

\vspace{0.5mm} \noindent \textbf{Textual feature extraction:}
We consider the transformer-based text encoder, BERT~\cite{devlin2018bert}.
A sentence is first tokenized into the token sequence, where the $[\mathtt{CLS}]$ and $[\mathtt{SEP}]$ tokens are added to the beginning and the end of the sequence.
The activation of the last layer of the transformer for the $[\mathtt{CLS}]$ token is employed as the textual feature $\mathbf{x}^{\mathcal{T}} \in \mathbb{R}^D$.

\subsection{Bottom-up Attention: Entity Discovery Module}
\vspace{-1mm}
\label{sec:slotattn}

{The entity discovery module applies a bottom-up attention mechanism to aggregate patch-based visual features into multiple entity embeddings that represent instances or object parts in the image.
The discovered entities are then treated as primitive units of the segmenation mask, where the mask is predicted by selectively combining them.
This approach is useful since the textual information is often built upon the relations between such visual entities.}

Specifically, the entity discovery module consists of two blocks: aggregation block and interaction block.
The \textit{entity slots}, which are embedding vectors that corresponds to various visual entities, are updated in an iterative process alternating between these two blocks. 
These slots eventually become entity embeddings.
Let $\Phi:\mathbb{R}^{N \times D}\mapsto \mathbb{R}^{K \times D}$ be the entity discovery module that takes visual features as input and outputs the set of $K$ entity embeddings.
The entity embeddings are obtained by iteratively updating initial entity slots $\mathbf{S}^0 \in \mathbb{R}^{K \times D}$ $T$-times:
{
\begin{align}
\begin{split}
    \Phi\big(\mathbf{x}^\mathcal{V}\big) & := \mathbf{S}^{T}, \ \ \textrm{where}\\
    \mathbf{S}^{t+1} &= \mathtt{Inter}\big(\mathtt{Agg}\big(\mathbf{x}^\mathcal{V}; \mathbf{S}^{t}\big)\big),
\label{eq:agg_and_prop}
\end{split}
\end{align}
and $\mathtt{Agg}$ and $\mathtt{Inter}$ denote the aggregation and interaction blocks, respectively.}

\begin{figure} [t!]
\centering
\includegraphics[width=0.93\columnwidth]{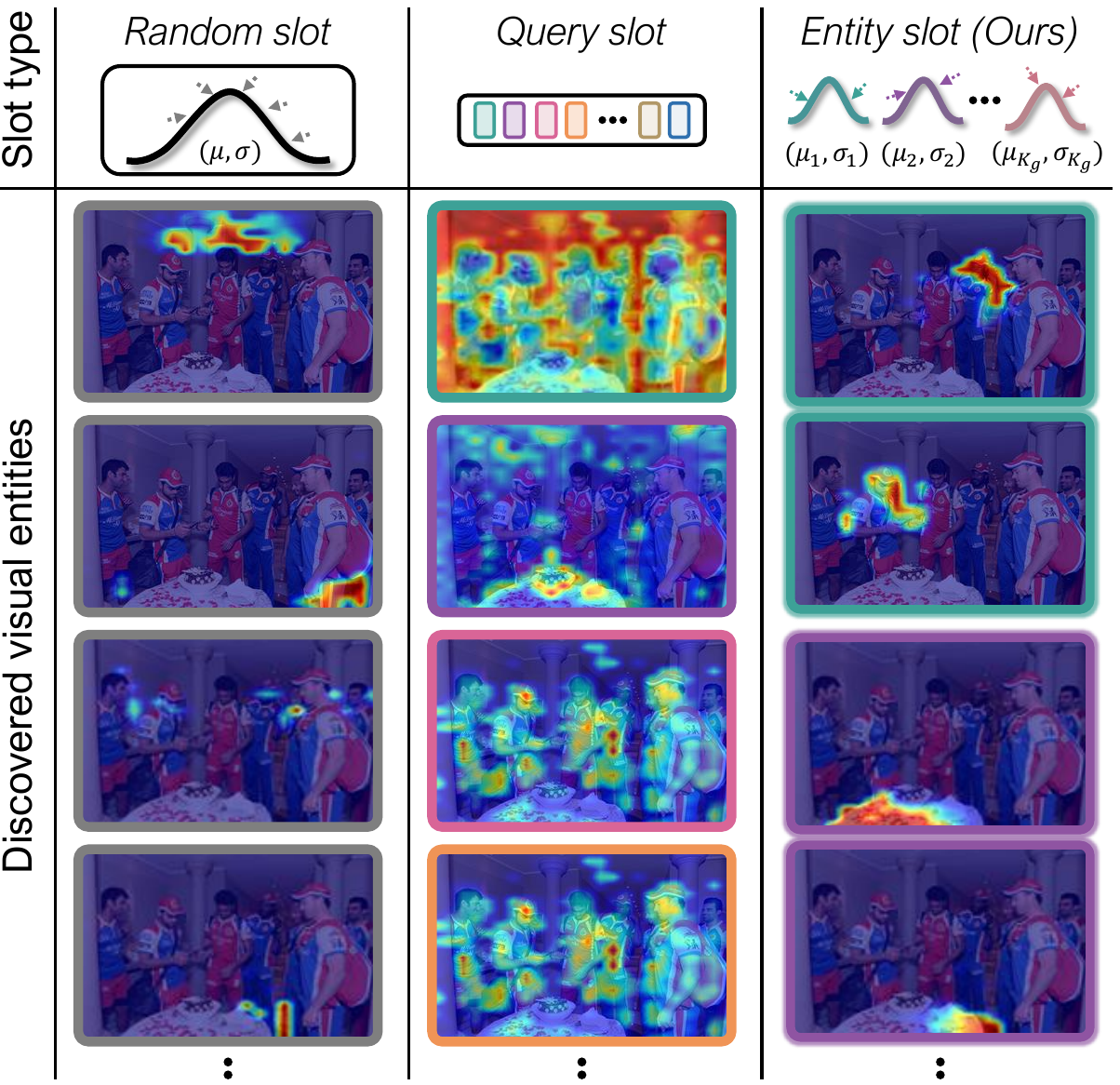}
\caption{
Comparison between the proposed entity slot and previous slot initialization methods.
The boundary color indicates the Gaussian distribution that each slot is sampled from.
The entity slot successfully discovers fine-grained entities such as head and food, while the others often fail to capture individual entities of specific semantics.
}
\vspace{-2.5mm}
\label{fig:slottype}
\end{figure}

In the literature, the initial slots $\mathbf{S}^0$ are often defined as either vectors sampled from a learnable Gaussian distribution~\cite{locatello2020object,singh2021illiterate} or learnable embeddings~\cite{xu2022groupvit,zhou2022slot,kim2022improving,Weinzaepfel2022LearningSF}, where such types of slots are referred as \textit{random slot} and \textit{query slot} in the remainder of this paper, respectively.
However, we have empirically found that both of these initialization strategies are unsuitable for our task as shown in~\Fig{slottype}.
{The random slot suffers from the lack of semantic specificity since every slot is randomly sampled from a single distribution.}
This can lead to suboptimal performance since entities in real-world images are diverse in terms of semantics.
The query slot, on the other hand, suffers from the limitation of being bound to specific semantic categories, such as human, animal, and vehicle.
This limitation leads to entity discovery that is not sufficiently fine-grained, \eg, 
{given an image of multiple people, a single slot that binds to ``human'' captures all of them and thus failed to identify the individual specified by the text query.
This is problematic particularly in our model since entities act as the smallest units for producing segmentation masks.}
{It is highly desired to have a mechanism that could guide the differentiation of slots to be semantically specific.}

To overcome these limitations, we introduce a new initialization strategy for the \emph{entity slot}, which enables the discovery of fine-grained visual entities while learning diverse semantics.
Formally, the initial entity slots $\mathbf{S}^0$ are obtained by sampling $K_s$ slots from each of $K_g$ different Gaussian distributions, resulting in a total of $K = K_s \cdot K_g$ slots:
\begin{align}
\begin{split}    
    \mathbf{S}^0 &:= [\mathbf{S}^0_{1}; ~\cdots;\mathbf{S}^0_{K_g}] \in \mathbb{R}^{K \times D},
    \ \ \text{where}
    \\
    \mathbf{S}^0_{i} &\sim \mathcal{N}(\mu_i, \mathtt{diag}(\sigma_i)) \in \mathbb{R}^{K_s \times D}.
    \label{eq:entityslots}
\end{split}
\end{align}
The idea is to initialize slots by sampling them from different Gaussian distributions, each specifying a distinct semantic property to be captured. 
Individual entities sharing the same semantics are differentiated by slots sampled from the same distribution, {which enables fine-grained entity discovery.}

From now, we present the aggregation block and interaction block at $t$-th iteration.
As described in \Eq{agg_and_prop}, the slots of previous iteration $\mathbf{S}^{t-1}$ are fed to the aggregation block with the visual features $\mathbf{x}^\mathcal{V}$. 
The role of the aggregation block is to let each slot aggregate visual features of distinct visual entities, and it is achieved by employing the competitive attention mechanism introduced in the slot attention~\cite{locatello2020object}.
Let $q(\cdot)$, $k(\cdot)$, and $v(\cdot)$ are linear projection with layer normalization~\cite{Ba2016LayerN} that map inputs to dimension $D_h$.
Then, the attention map $A^{\text{slot}} \in \mathbb{R}^{N \times K}$ between $\mathbf{x}^\mathcal{V}$ and $\mathbf{S}^{t-1}$ is computed as follows:
\begin{align}
\small
    A^{\text{slot}}_{i,j} = \frac{
        e^{M_{i,j}}
    }{
        \sum_{j=1}^{K} e^{M_{i,j}}
    },\text{~where~}
    M = \frac{
        k\big(\mathbf{x}^\mathcal{V}\big) q\big(\mathbf{S}^{t-1}\big)^{\top}
    }{
        \sqrt{D_h}
    }.
    \label{eq:aslot}
\end{align}
Unlike the cross-attention transformer~\cite{vaswani2017attention} that normalized attention over the \textit{keys}, we normalize attention map $A^{\text{slot}}$ over the \textit{slots}.
This normalization scheme encourages slots to compete for aggregating the visual features, which results in the set of slots binding to different visual entities in the input data.
It is worth noting that $A^{\text{slot}}$ provides the localizations of entities discovered, which is used as primitive units of the segmentation mask during inference.
Using $A^{\text{slot}}$, the slots are updated in a similar manner to the transformer:
\begin{align}\begin{split}
    \mathbf{\widehat{S}^{t}} &= \mathtt{Agg}\big(\mathbf{x}^\mathcal{V}; \mathbf{S}^{t}\big)\\
    &={\mathtt{normalize}}(
        A^{\text{slot}}
    )^{\top} v\big(\mathbf{x}^\mathcal{V}\big) W_o + \mathbf{S}^{t-1}, 
\end{split}\end{align}
where $\mathtt{normalize}(\cdot)$ denotes $\ell_1$-normalization of the column vectors and $W_o \in \mathbb{R}^{D_h \times D}$ is linear projection.
Finally, slots are further refined by $~\widehat{\mathbf{S}}^{t} \gets \mathtt{MLP}\big(~\widehat{\mathbf{S}}^{t}\big) + ~\widehat{\mathbf{S}}^{t}$, where $\mathtt{MLP}$ is multi-layer perceptron (MLP) with layer normalization.

The following interaction block consists of multiple self-attention transformers~\cite{vaswani2017attention}. 
The weights of these transformers are shared, and each transformer is applied to slots sampled from the same distribution.
We denote the transformer layer as $\mathtt{Transformer(Q; K; V)}$, where $\mathtt{(Q; K; V)}$ are for query, key, and value.
The interaction block is denoted as follows:
\begin{align}
\small{
    \mathtt{Inter}\big(\mathbf{\widehat{S}}^{t}\big) = [\mathtt{SA}\big(\mathbf{\widehat{S}}^{t}_1\big);\cdots; \mathtt{SA}\big(\mathbf{\widehat{S}}^{t}_{K_g}\big)],
}
\end{align}
where $\mathtt{SA(X)}$ is for the $\mathtt{Transformer(X; X; X)}$.
This self-attention scheme models information exchange between visual entities sharing the same semantic property, which helps slots within a single semantic category to capture different entities without redundancy.

\subsection{Top-down Attention: Modality Fusion Module}
\vspace{-1mm}
\label{sec:crossattn}
The modality fusion module is proposed for top-down attention mechanism, which combines discovered visual entities based on the textual feature.
Let $\Psi: \mathbb{R}^{K \times D} \times \mathbb{R}^{D} \mapsto \mathbb{R}^{D}$ be the modality fusion module, which takes entity embeddings $\Phi(\mathbf{x}^\mathcal{V})$ and textual feture $\mathbf{x}^\mathcal{T}$ as inputs. 
Then, the output cross-modal embedding $\mathbf{z} \in \mathbb{R}^{D}$ is computed using cross-attention transformer:
\begin{align}
    \mathbf{z} = \Psi\big(\Phi(\mathbf{x}^\mathcal{V}); \mathbf{x}^\mathcal{T}\big)
    = \frac{\mathtt{CA}\big(\Phi(\mathbf{x}^\mathcal{V}); \mathbf{x}^\mathcal{T}\big)}{\lVert\mathtt{CA}\big(\Phi(\mathbf{x}^\mathcal{V}); \mathbf{x}^\mathcal{T}\big)\rVert_{2}},
\end{align}
where $\mathtt{CA(X;Y)}$ is for the $\mathtt{Transformer(Y;X;X)}$ and $\lVert\cdot\rVert_2$ denotes $\ell_2$ norm.
Moreover, we can obtain the attention map $A^{\text{fuse}} \in \mathbb{R}^{K}$ as a byproduct in cross-attention:
\begin{align}
    A^{\text{fuse}}_{i} = \frac{
        e^{M_{i}}
    }{
        \sum_{i=1}^{K} e^{M_{i}}
    },\text{~where~}
    M = \frac{
        k\big(\Phi(\mathbf{x}^\mathcal{V})\big) q\big(\mathbf{x}^\mathcal{T}\big)
    }{
        \sqrt{D_h}
    }.
    \label{eq:afuse}
\end{align}
Using cross-attention, the module effectively reasons the relevance between entities and the textual feature.
Inferred relevance is described in the $A^{\text{fuse}}$, which is the probability of each entity being referred to in the textual feature.

\subsection{Training}
\vspace{-1mm}
\label{sec:train}
Our model is trained with two different loss functions: contrastive cycle-consistency (C$^3$) and reconstruction loss.

\vspace{-2mm}
\subsubsection{Contrastive cycle-consistency (C$^3$) Loss}
\vspace{-2mm}
The output of our model is the cross-modal embedding, which is obtained by fusing information from visual and textual modalities.
The fusion should selectively combine discovered entities based on textual information, but a segmentation label for which entities are referred to is not given in our weakly-supervised setting.
To resolve this issue, we add an additional constraint to the contrastive learning objective: establishing \textit{cycle-consistency} between each ground-truth pair of image and text.

Let $\{(\mathbf{x}_i^\mathcal{V}, \mathbf{x}_i^\mathcal{T})\}_{i=1}^{B}$ represent a batch of visual and textual features, and $\mathbf{z}_{ij}= \Psi\big(\Phi(\mathbf{x}_i^\mathcal{V}); \mathbf{x}_j^\mathcal{T}\big)$ denote the cross-modal embedding produced with $\mathbf{x}_i^\mathcal{V}$ and $\mathbf{x}_j^\mathcal{T}$. 
For simplicity, we refer to the cross-modal embedding produced using the ground-truth image-text pair as matching embedding, and all other cross-modal embeddings as non-matching embeddings.
Then C$^{3}$ loss is computed as follows:
\begin{align}
\begin{split}
    \mathcal{L}_{\text{C}^3} &= -\frac{1}{B} \sum_{j=1}^B \log\frac{e^{\langle \mathbf{z}_{jj}, \mathtt{sg}(\mathbf{x}_j^\mathcal{T}) \rangle}}{\sum_{i=1}^B e^{\langle \mathbf{z}_{ij}, \mathtt{sg}(\mathbf{x}_j^\mathcal{T}) \rangle}}
\end{split}
\end{align}
where $\langle\cdot,\cdot\rangle$ and $\mathtt{sg}(\cdot)$ denote inner product and stop-gradient operation, respectively.
Stop-gradient operation is used to prevent the collapsing solutions where every embedding becomes similar, as discussed in~\cite{chen2021exploring}.

The proposed loss function encourages the matching embeddings to be similar to the input textual feature, while making the non-matching embeddings to be dissimilar.
This approach ensures that the modality fusion module preserves the inherent information of the referring expression only when it is paired with the matching image (cycle-consistency), achieved by attending to the referred entities that may be absent in non-matching images.
Therefore, an attention map $A^{\text{fuse}}$ is trained to reflect the referring expression accurately, where relevance between the expression and entities automatically emerges.

\vspace{-2mm}
\subsubsection{Reconstruction Loss}
\vspace{-2mm}
If we can reconstruct the original data using only the set of slots, then the slots will correspond to distinctive individual entities in the input image.
Therefore, we leverage the reconstruction objective $\mathcal{L}_{\text{recon}}$, which is given by
\begin{align}
\small
    \mathcal{L}_{\text{recon}} = \frac{1}{B} \sum_{i=0}^B \big\lVert \mathtt{sg}\big(\mathbf{x}_i^\mathcal{V}\big) - f_{\text{dec}}\big(\Phi\big(\mathbf{x}_i^\mathcal{V}\big)\big) \big\rVert^2,
\end{align}
where $f_\text{dec}(\cdot)$ is a MLP based decoder.
Unlike the original slot attention~\cite{locatello2020object}, our reconstruction is performed in the feature space, rather than the pixel space.
This feature reconstruction scheme enables an efficient decoder design.
Detailed descriptions are provided in the supplementary material.

\subsection{Inference}
\vspace{-1mm}
\label{sec:infer}
The segmentation mask is computed with the two attention maps: $A^\text{slot} \in \mathbb{R}^{N \times K}$ and $A^\text{fuse} \in \mathbb{R}^{K}$.
$A^\text{slot}$ is patch-wise attention map that contains information about the discovered entities (\Eq{aslot}), while $A^\text{fuse}$ is entitiy-wise attention map denoting relevance between the referring expression and the entities (\Eq{afuse}).
To compute the segmentation mask, we first compute $A^\text{slot} \cdot A^\text{fuse}$, which results in a 1-dimensional vector of size $N$. 
We then reshape this vector into a 2-dimensional map. 
Next, we bilinearly interpolate the map to the original image size and rescale it using min-max normalization. 
Finally, we binarize the map using a threshold value $\tau$ to obtain the segmentation mask.

\begin{table*}[t!]
\setlength{\tabcolsep}{4pt}
\centering
\scalebox{0.85}{
\begin{tabular}{p{2.2cm}<{}|p{3.6cm}<{\centering}|p{1cm}<{\centering}|p{1cm}<{\centering}p{1cm}<{\centering}p{1cm}<{\centering}|p{1cm}<{\centering}p{1cm}<{\centering}p{1cm}<{\centering}|p{1cm}<{\centering}|p{1.5cm}<{\centering}}
\toprule
\multirow{2}{*}{Methods} 
& \multirow{1}{*}{Image-text}
& \multirow{1}{*}{Fine}
& \multicolumn{3}{c|}{RefCOCO}
&\multicolumn{3}{c|}{RefCOCO+} 
&\multicolumn{1}{c|}{Gref}
&\multicolumn{1}{c}{PhraseCut} \\
                                                    &  pretraining dataset  & tuning & \textit{val} & \textit{testA} & \textit{testB}   & \textit{val} & \textit{testA} & \textit{testB}  & \textit{val} & \textit{val} \\ \midrule
\multirow{2}{*}{GroupViT~\cite{xu2022groupvit}}     & \multirow{2}{*}{CC12M+YFCC (26M)}           & \xmark &7.99   & 6.16  & 10.51 & 8.49  & 6.79  & 10.59 & 10.68 & 5.82 \\
                                                    &                                       & \cmark & 10.82 & 11.11 & 11.29 & 11.14 & 10.78 & 11.84 & 12.77 & 9.41 \\ \midrule 
\multirow{2}{*}{MaskCLIP~\cite{zhou2022extract}}    & \multirow{2}{*}{WIT (400M)}      & \xmark & 11.52 & 11.85 & 12.06 & 11.87 & 12.01 & 12.57 & 12.74 & 12.13\\ 
                                                    &                                       & \cmark & 19.45 & 18.69 & 21.37 & 19.97 & 18.93 & 21.48 & 21.11 & 23.80 \\ \midrule %
TSEG~\cite{strudel2022weakly}                            &   -                   & \cmark & 25.44 & - & - & 22.01 & - & - & 22.05 & 28.77\\ \midrule
\multirow{2}{*}{Ours}                               & Visual Genome (0.31M)  & \xmark & 21.80 & 19.00 & 24.96 & 22.20 & 19.86 & 24.85 & 25.89 & - \\
                                                    & -  & \cmark & \textbf{34.76} & \textbf{34.58} & \textbf{35.01} & \textbf{28.48} & \textbf{28.60} & \textbf{27.98} & \textbf{28.87}  & \textbf{33.45} \\
\bottomrule
\end{tabular}
}
\vspace{-1mm}
\caption{
Comparison with weakly-supervised method (TSEG) and open-vocabulary segmentation method (GroupViT and MaskCLIP). The results on four datasets are reported in mIoU (\%). Fine-tuning \cmark means that the model is trained with the image-text pairs of the target benchmark; otherwise, it just relies on the pre-trained weight without further training.
}
\vspace{-2mm}
\label{tab:comp_open}
\end{table*}

\vspace{-1mm}
\section{Experiments}
\vspace{-0.5mm}
\label{sec:experiment}

\subsection{Experimental Setting}
\vspace{-1.5mm}
\noindent \textbf{Datasets:}
Our experiments utilize four datasets commonly used in referring image segmentation task: RefCOCO~\cite{yu2016modeling}, RefCOCO+~\cite{yu2016modeling}, Gref~\cite{mao2016generation}, and PhraseCut~\cite{wu2020phrasecut}.
RefCOCO, RefCOCO+, and Gref are obtained from COCO~\cite{Mscoco} dataset. RefCOCO contains 142,209 image-text pairs, while RefCOCO+ has 141,564 pairs and does not include expressions indicating orientation properties (\eg, top, right).
Gref includes 104,560 pairs for 54,822 objects.
PhraseCut comprises 345,486 image-text pairs, which are derived from Visual Genome~\cite{visual_genome}.

\noindent \textbf{Implementation details:}
For visual and textual backbone, we employ ViT-S-16 pretrained on ImageNet-21K~\cite{Imagenet} and pretrained BERT~\cite{devlin2018bert}, respectively.
We make each modality feature have the same channel dimension $D$ by applying a 1$\times$1 convolution layer.
We set $D$ and $D_h$ to 512 and 1024, respectively.
In the aggregation block, the number of iteration $T$ is set to 6, and the number of slots $K$ is set to 36 except for PhraseCut, which is set to 60.
For the entity slot, the number of slots sampled from each Gaussian distributions $K_s$ is set to 3 in RefCOCO+ and Gref, while 2 in RefCOCO and PhraseCut.
During training, we use AdamW~\cite{loshchilov2017decoupled} optimizer, with the initial learning rate of 1e-4 scheduled with cosine annealing~\cite{loshchilov2016sgdr}.
Batch size is set to 32 and model is trained for 50 epochs.
We resize input images to 384 $\times$ 384 for both training and evaluation.
Threshold value $\tau$ is set to $0.5$ for all experiments.

\noindent \textbf{Evaluation metric:}
Following~\cite{yang2022lavt}, we employ the cumulative intersection-over-union (cIoU) and mean intersection-over-Union (mIoU) metric;
cIoU divides the total intersections by the total unions across all test images, and mIoU computes IoU for each image and averages them.
Also, we evaluate the accuracy at the IoU thresholds of \{0.3, 0.5, 0.7\}, denote as $\mathcal{A}$@\{0.3, 0.5, 0.7\} in Table~\ref{tab:slot_var}.

\subsection{Comparisons with the State of the Art}
\vspace{-1.5mm}

\noindent \textbf{Weakly supervised methods:}
As presented in~\Tbl{comp_open}, our method outperforms TSEG~\cite{strudel2022weakly}, the previous work on weakly-supervised referring image segmentation, in all benchmarks.
It shows that the proposed bottom-up and top-down attention framework is more effective than using patch-based visual features with a carefully designed similarity function of TSEG.

\noindent \textbf{Open-vocabulary segmentation:}
In~\Tbl{comp_open}, we also compare our model with the recent open-sourced open-vocabulary segmentation models: GroupViT~\cite{xu2022groupvit} and MaskCLIP~\cite{zhou2022extract}, on the referring image segmentation benchmarks. Note that these methods are pre-trained with large-scale image-text datasets~\cite{changpinyo2021conceptual, thomee2016yfcc100m, radford2021learning}.
We also report their fine-tuned version for fair comparisons: models are additionally trained with the image-text pairs of each benchmark. Further details for the reproductions are provided in the supplementary material.
Ours outperforms GroupViT and MaskCLIP in all settings and benchmarks, even without large-scale image-text pretraining. This result shows that open-vocabulary segmentation models~\cite{zhou2022extract, xu2022groupvit} have difficulty comprehending the complex free-from text involving the relation between entities, unlike ours.
Interstingly, our model without fine-tuning, which pre-trained on the PhraseCut dataset and evaluated on other benchmarks, still achieved better results.
This setting is identical to the open-vocabulary segmentation models, except it leverages a far smaller dataset.
These results demonstrate the feasibility of our method in a zero-shot transfer setting.

\begin{figure*}[t!]
    \centering
    \includegraphics[width=\linewidth]{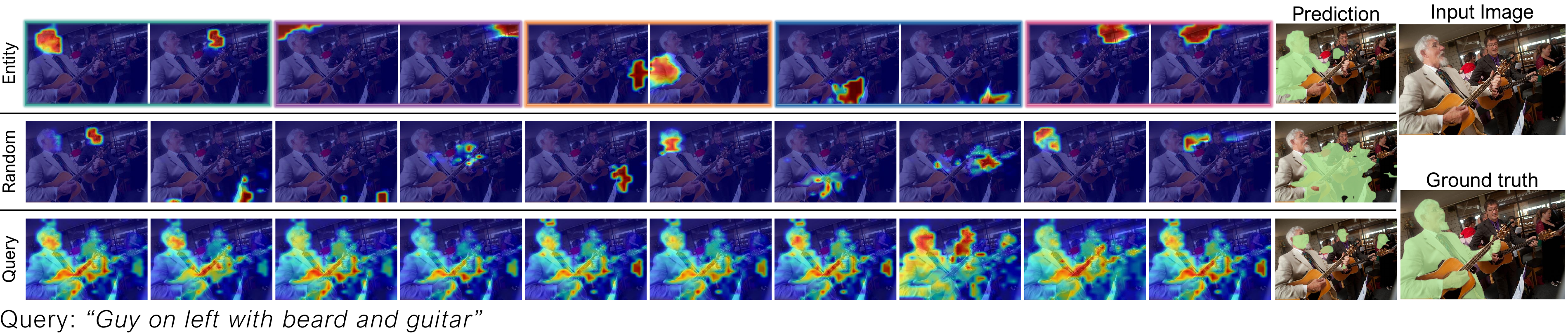}
\vspace{-5mm}
\caption{
Qualitative results of our framework with entity slot (Ours), random slot, and query slot on RefCOCO \textit{val} set. 
In the case of entity slots, the boundary color indicates the Gaussian distribution that each slot is sampled from.
For each slot type, we present 10 discovered entities from $A^\text{slot}$ and the final predictions.
}
\vspace{-2mm}
\label{fig:slotqual}
\end{figure*}

\subsection{Ablation Study}
\vspace{-1.5mm}
We perform ablation studies on the \textit{val} set of RefCOCO to analyze the contributions of the entity discovery module and the modality fusion module.

\vspace{0.5mm} \noindent \textbf{Entity discovery module:}
In~\Tbl{ablation}(a), we ablate the entity discovery module and use visual feature directly, which leads to the drastical performance drop. This result demonstrates that contribution of our bottom-up attention framework is significant.
Moreover, we ablate the interaction block, replace slot-attention of aggregation block to transformer, and train the model without reconstruction loss.
The results show that all of the design choices for the module is important for effective entity discovery.

\vspace{0.5mm} \noindent \textbf{Modality fusion module:}
In~\Tbl{ablation}(b), the modality fusion module is ablated, and the model is trained with image-text contrastive learning~\cite{radford2021learning}, where the image feature is obtained by pooling the entity embeddings.
The result demonstrates that inferring relevance between entities and referring expression with modality fusion module, trained with C$^3$ loss, is crucial for accurate segmentation.

\begin{table}[t!]
\centering
\scalebox{0.9}{
\begin{tabular}{l|cc}
\toprule
{}                                        & cIoU & mIoU                    \\ \midrule
\multicolumn{3}{l}{(a) \textit{\textbf{Analysis on entity discovery module}}}  \\ \midrule
{Ours}                    & \textbf{30.40} & \textbf{34.76}         \\
{w/o entity discovery module} & 5.40 &  5.91                   \\ 
{w/o interaction block}                   & 29.59 & 33.69                  \\
{aggregation block w/ transfomer}                   & 11.29 & 12.61                 \\
{w/o reconstruction loss}                              & 18.61 &  21.84         \\ \midrule
\multicolumn{3}{l}{(b) \textit{\textbf{Analysis on modality fusion module}}}     \\ \midrule
{Ours}        & \textbf{30.40} & \textbf{34.76}         \\
{w/o modality fusion module and C$^3$ loss}  & 13.81 & 14.65                  \\  \bottomrule
\end{tabular}
}
\vspace{-1mm}
\caption{Ablation studies of model components and loss function on RefCOCO \textit{val} set. 
}
\vspace{-3mm}
\label{tab:ablation}
\end{table}

\begin{figure*}[t!]
    \centering
    \includegraphics[width=\linewidth]{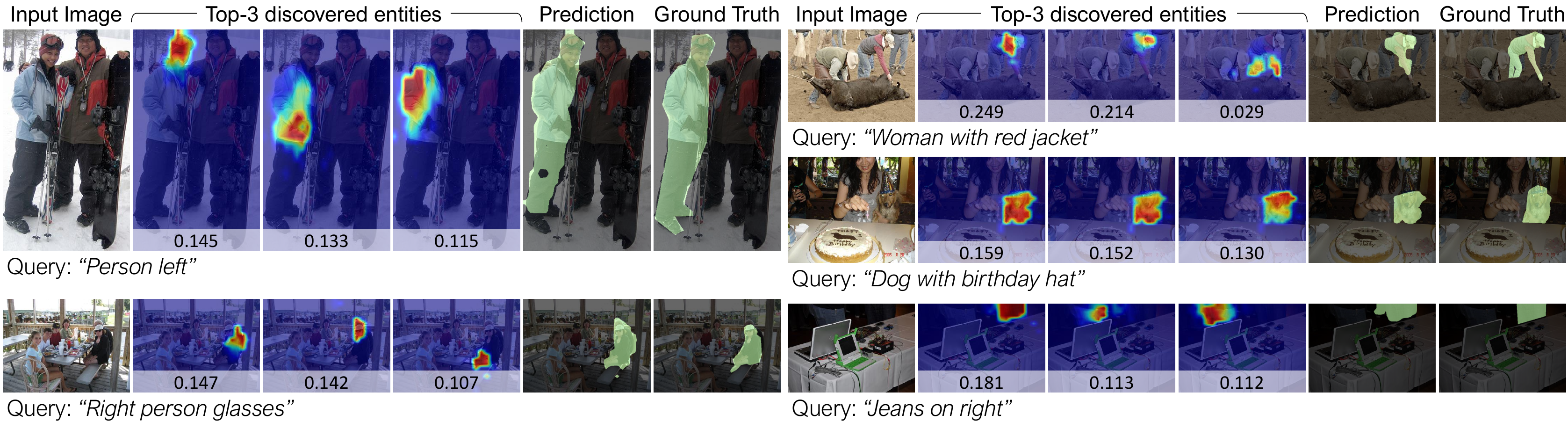}
\vspace{-5.5mm}
\caption{
Qualitative results of our framework on RefCOCO \textit{val} set.
We present the discovered entities from $A^\text{slot}$ and their relevance scores from $A^\text{slot}$. Top-3 entities in terms of relevance to query expression are presented.
} 
\vspace{-3mm}
\label{fig:segqual}
\end{figure*}

\subsection{In-depth Analysis}
\vspace{-1.5mm}

\noindent \textbf{Analysis of slot initialization:}
To verify our design choice for entity slots, we investigate variants of slots; query, random, and entity slot.
As summarized in~\Tbl{slot_var}, our entity slot outperforms other slot types across the different numbers of slots.
Specifically, as the number of slots is increased, the performance of the query slot declines, and that of the random slot converges.
This result implies that simply increasing the number of slots can not handle the discovery of fine-grained visual entities in real-world images, and thus gains from the proposed entity slot type are non-trivial.
In~\Fig{slotqual}, we also present several visual entities discovered using each slot type, which show the superiority of the entity slot against others as explained in~\Sec{slotattn}.
Specifically, the entity slots from the same distribution share the same semantic property (\eg, head, body, or window) while capturing different fine-grained entities.
The proposed entity slots effectively discover fine-grained visual entities, which results in an accurate prediction.
In contrast, the prediction using random slots leads to a noisy mask, and using query slots leads to the mask of every head that appears in the image, respectively.

\begin{table}[t!]
\centering
\scalebox{0.8}{
\begin{tabular}{l|c|ccc|cc}
\toprule
Slot type & \textit{\# slots} & \multicolumn{1}{c}{$\mathcal{A}$@0.3} & \multicolumn{1}{c}{$\mathcal{A}$@0.5} & \multicolumn{1}{c|}{$\mathcal{A}$@0.7} & \multicolumn{1}{c}{cIoU} & mIoU \\ \midrule
\multirow{3}{*}{Query} & 12 & 27.57	&  7.17 & 0.96 & 22.04 & 21.44 \\
                       & 24 & 36.73 & 10.61 & 1.93	& \textbf{23.59} & \textbf{24.78} \\
                       & 36 & 19.30 &  3.48 & 0.31	& 16.66 & 18.16 \\ \midrule
\multirow{3}{*}{Random}  & 12 & 36.30 &	11.27 &	2.69 & 24.20 & 26.65 \\
                           & 24 & 45.19 & 16.01 & 3.66 & \textbf{26.60} & 29.87 \\
                           & 36 & 45.62 & 16.88 & 3.76 & 26.55 & \textbf{30.15} \\ \midrule
\multirow{3}{*}{\shortstack{Entity \\ (Ours)}} & 12 & 39.43 & 13.87 & 3.56 & 24.72 & 27.84 \\
                           & 24 & 52.33 & 22.97 & 6.23 & 29.22 & 33.54 \\
                           & 36 & 55.02 & 24.99 & 6.35	& \textbf{30.40} & \textbf{34.76}\\
\bottomrule
\end{tabular}
}
\vspace{-1.5mm}
\caption{
Comparison between different slot types on RefCOCO \textit{val} set, where \textit{\# slots} is the total number of slots.
}
\vspace{-3mm}
\label{tab:slot_var}
\end{table}

\noindent \textbf{Effect of hyperparameters $K_g$ and $K_s$:}
In~\Tbl{kgks_afuse}(a), we investigate the impact of hyperparameters for the entity slots: $K_g$ and $K_s$, which are the number of Gaussian distributions and slots sampled from each distribution, respectively. 
We keep the total number of entity slots to 36, only changing $K_g$ and $K_s$.
The optimal configurations are 18 and 2 for $K_g$ and $K_s$, respectively.
Interestingly, we observed that even entity slots with $K_s=1$, where a single slot is sampled from each distribution, vastly surpass the query slots in~\Tbl{slot_var}; it shows that random sampling of slots prevents the bounding to specific semantic category.

\noindent \textbf{Effectiveness of $A^{\text{fuse}}$:}
To verify the efficacy of $A^\text{fuse}$ quantitatively,
we evaluate our model with the following inference scheme variants:
(1) $\mathtt{AVG}$ that replaces $A^\text{fuse}$ with the uniform probability distribution,
(2) $\mathtt{MAX}$ that picks a single attention map from $A^\text{slot}$ with the maximum probability in $A^\text{fuse}$, and
(3) $\mathtt{MIN}$ that picks the attention map with the minimum probability.
As summarized in~\Tbl{kgks_afuse}(b), $\mathtt{MAX}$ largely outperforms $\mathtt{AVG}$ even though it only uses a single attention map for the final prediction, while $\mathtt{MIN}$ results in a near-zero cIoU and mIoU.
This tendency shows that $A^\text{fuse}$ effectively assigns probability based on the relevance between reference and entities, even without direct supervision.

\begin{table}[t!]
    \centering
    \begin{subtable}[h]{0.194\textwidth}
        \centering
        \scalebox{0.77}{
        \begin{tabular}{cc|ccc|cc}
        \toprule
        $K_g$  & $K_s$ & cIoU & mIoU              \\ \midrule
        36 & 1 & 29.36 & 33.41            \\
        18 & 2 & \textbf{30.40} & \textbf{34.76}   \\
        12 & 3 & 29.63 & 34.03            \\
        9  & 4 & 28.31 & 32.34                \\
        6  & 6 & 27.23 & 31.30                \\
        \bottomrule
        \end{tabular}
        }
        \vspace{-2mm}
        \caption{}
        \label{tab:attn_a}
     \end{subtable}
    \hfill
    \begin{subtable}[h]{0.278\textwidth}
        \centering
        \small
        \scalebox{0.88}{
        \begin{tabular}{l|cc}
        \toprule
        Inference scheme   & cIoU       & mIoU      \\ \midrule
        $\mathtt{AVG}$     &   9.01     & 8.16          \\
        $\mathtt{MAX}$     &   14.93    & 19.54     \\
        $\mathtt{MIN}$     &   0.16     & 0.14          \\ \midrule
        $A^{\text{fuse}}\cdot A^{\text{slot}}$ (Ours)  & \textbf{30.40} & \textbf{34.76} \\
        \bottomrule
        \end{tabular}
      }
        \vspace{-2mm}
        \caption{}
      \label{tab:variantencoder}
    \end{subtable}
    \vspace{-1.5mm}
    \caption{
    Impact of hyperparameters $K_g$ and $K_s$ (a) and inference schemes (b) on RefCOCO \textit{val} set.
    }
    \vspace{-3mm}
     \label{tab:kgks_afuse}
\end{table}

\noindent \textbf{Qualitative Analysis:}
In~\Fig{segqual}, predicted segmentation mask and discovered entities with top-3 relevance are presented.
Results show that the entity discovery module effectively discovers primitive units of segmentation masks, and the modality fusion module accurately infers their relevance to query referring expression.

\vspace{-1mm}
\section{Conclusion} 
\vspace{-1.5mm}
In this paper, we address the problem of learning referring image segmentation with only image-text pairs.
We present a new model based on a bottom-up and top-down attention framework, which first discovers entities in the input image and then combines such entities based on their relevance to the text query.
Moreover, we present a new loss function enforcing cycle-consistency between image-text pairs, which enables the effective training of the model without any further supervision.
As a result, the proposed method clearly outperforms an existing weakly-supervised method and recent open-vocabulary segmentation models.
For future work, we will explore expanding our model to different modalities such as video and audio.
\label{sec:conclusion}

\vspace{2mm}
{\small
\noindent \textbf{Acknowledgement.} 
This work was supported by 
the NRF grant and  %
the IITP grants     %
funded by Ministry of Science and ICT, Korea
(NRF-2021R1A2C3012728--20\%,  %
 IITP-2020-0-00842--20\%,     %
 IITP-2022-0-00290--20\%,     %
 IITP-2022-0-00926--20\%,     %
 IITP-2021-0-02068--10\%,     %
 IITP-2019-0-01906--10\%).    %
}

\pagebreak

{\small
\bibliographystyle{ieee_fullname}
\bibliography{cvlab_kwak}
}

\appendix

\twocolumn[
  \begin{@twocolumnfalse}
    \huge{\textbf{Appendix}}
    \vspace{5mm}
  \end{@twocolumnfalse}
]

\addcontentsline{toc}{section}{Appendices}
\renewcommand\thefigure{A\arabic{figure}}
\renewcommand{\thetable}{A\arabic{table}}
\setcounter{figure}{0}
\setcounter{table}{0}

\normalsize{
This supplementary material presents experimental results omitted from the main paper due to the space limit.
We first summarize the notations in our paper in~\Tbl{notation}.
\Sec{supp_analy_iter} analyzes performance according to the number of iterations of the entity discovery module $T$ and thresholding value $\tau$.
In~\Sec{supp_exp_detail}, we describe experimental details of reproducing open-vocabulary segmentation methods~\cite{xu2022groupvit, zhou2022extract}.
\Sec{supp_decoder} provides the details of the decoder used in reconstruction loss.
We then present quantitative results with the DenseCRF~\cite{Fullycrf} in~\Sec{supp_more_quan}.
Finally, \Sec{supp_more_qual} offers more qualitative results of our model on the RefCOCO \textit{val} set.
Since our model involves randomness when sampling the entity slots, all of the results in the main paper and supplementary material are obtained by averaging the results from 3 experiments, where the standard deviation across them is always less than 0.06.

\newcommand*{\mytab}{\hspace*{0.35cm}}

\begin{table}[t!]
\scalebox{0.71}{
    \begin{tabular}{p{1.85cm}<{}p{9.1cm}<{}}
    \toprule
    Symbol        & Description~~~~~~~~~~~~~~~~~~~~~~~~~~~~~~~~~~~~     \\
    \midrule\multicolumn{2}{l}{\textit{\textbf{Feature extraction}}} \\
    $\mathbf{x}^\mathcal{V}$ & The visual feature \\
    $\mathbf{x}^\mathcal{T}$ & The textual feature \\
    
    \midrule\multicolumn{2}{l}{\textit{\textbf{Entity discovery module}}} \\
    $\Phi$                                  & The entity discovery module        \\   
    $\mathtt{Agg}$            & The aggregation block        \\
    \mytab$\mathtt{normalize}$        & $\ell_1$ normalization of column vectors        \\   
    \mytab$\mathtt{MLP}$    & A multi-layer perceptron with layer normalization       \\
    \mytab$\mathbf{S}^0$          & Initial slots        \\
    \mytab$\mathbf{S}^t$          & Slots at $t$-th iteration          \\   
    \mytab$\hat{\mathbf{S}}^t$      & Output slots of the aggregation block at $t$-th iteration      \\
    \mytab${\mathbf{S}}_i^t$        & Slots sampled from the $i$-th distribution at $t$-th iteration \\
    $\mathtt{Inter}$          & The interaction block        \\   
    \mytab$\mathtt{SA}$ & A self-attention transformer        \\
    
    \midrule\multicolumn{2}{l}{\textit{\textbf{Modality fusion module}}} \\
    $\Psi$                                  & The modality fusion module       \\   
    $\mathtt{CA}$                           & A cross-attention transformer        \\   
    
    \midrule\multicolumn{2}{l}{\textit{\textbf{Training}}} \\
    $\langle\cdot,\cdot\rangle$                    & A inner product between two vectors      \\
    $\mathtt{sg}$                                  & A stop-gradient operation        \\   
    $f_\text{dec}$                                 & The reconstruction decoder        \\

    \midrule \multicolumn{2}{l}{\textit{\textbf{Inference}}} \\
    $A^\text{slot}$  & The patch-wise attention map from the entity discover module      \\   
    $A^\text{fuse}$  & The entity-wise attention map from the modality fusion module       \\   
    
    \midrule\multicolumn{2}{l}{\textit{\textbf{Hyperparameters}}} \\
    $N$   & The number of image patches        \\   
    $K$   & The total number of slots        \\
    $K_g$   & The number of different Gaussian distributions for entity slot        \\   
    $K_s$   & The number of slots sampled from each distribution for entity slot        \\   
    $D$, $D_h$   & The output and hidden dimension \\   
    $T$   & The number of iterations in entity discovery module        \\   
    $\tau$   & The threshold for predicting mask        \\  \bottomrule
    \end{tabular}
}
\vspace{-2mm}
\caption{
Summary of notations used in the main paper together with descriptions.
}
\vspace{-4mm}
\label{tab:notation}
\end{table}

\begin{table}[t!]
\centering
\scalebox{1}{
\begin{tabular}{c|ccc|cc}
\toprule
$T$   & $\mathcal{A}$@0.3 & $\mathcal{A}$@0.5 & $\mathcal{A}$@0.7 & cIoU & mIoU            \\ \midrule\midrule
8       & 53.08	& 23.30	& 5.98 & 29.38 & 33.85                  \\
\ccol 6 & \ccol 55.02	& \ccol 24.99	& \ccol 6.35 & \ccol \textbf{30.40} & \ccol \textbf{34.76}         \\
4       & 52.03	& 22.55	& 5.85 & 28.95 & 33.24                  \\
2       & 48.40	& 19.92	& 5.09 & 27.67 & 31.77                  \\
1       & 40.36	& 12.93	& 2.73 & 25.11 & 28.00                  \\  \bottomrule
\end{tabular}
}
\vspace{-1mm}
\caption{
Performance analysis according to the number of iteration $T$ of the entity discovery module on RefCOCO \textit{val} set.
The setting used in the main paper is indicated with a grey-colored row.
}
\vspace{-1mm}
\label{tab:iteration}
\end{table}

\section{Impact of hyperparameters $T$ and $\tau$}
\label{sec:supp_analy_iter}
\noindent\textbf{The number of iterations $T$:}
In~\Tbl{iteration}, We investigate the impact of the number of iterations in the entity discovery module $T$.
The model consistently achieves high IoU and accuracy when $T$ is greater than 2.
Notably, we observe a significant performance degradation when $T$ equals 1 (3.77\%p mIoU drop compared to when $T$ is 2).
These results highlight the significance of iteratively applying the aggregation and interaction blocks, which enables effective visual entity discovery through the progressive refinement of slots.
Furthermore, it is worth noting that even the least-performing model with a single iteration still outperforms the previous work~\cite{strudel2022weakly, xu2022groupvit, zhou2022extract}.
\begin{table}[t!]
\centering
\scalebox{0.9}{
\begin{tabular}{p{1.2cm}<{}|p{1cm}<{\centering}p{1cm}<{\centering}p{1cm}<{\centering}p{1cm}<{\centering}p{1cm}<{\centering}}
\toprule
$\tau$  &  0.3  &  0.4  &  0.5  &  0.6  &  0.7  \\ \midrule
mIoU    & 32.58 & 34.39 & \textbf{34.76} & 33.57 & 30.72 \\ \bottomrule
\end{tabular}
}
\vspace{-1mm}
\caption{
Performance analysis according to thresholding hyperparameter $\tau$ on RefCOCO \textit{val} set.
}
\vspace{-4mm}
\label{tab:threshold}
\end{table}

\vspace{1mm} \noindent \textbf{{The threshold value $\tau$:}}
In~\Tbl{threshold}, we present the mIoU of predictions from our model with the varying threshold value $\tau$.
Our model consistently attains high IoUs when $\tau$ ranges between 0.4 and 0.6, which is why we set $\tau$ to 0.5 for all experiments in the main paper. These findings indicate that our model is insensitive to the setting of $\tau$.

\section{Reproducing MaskCLIP and GroupViT}
\label{sec:supp_exp_detail}
To evaluate referring image segmentation performance of the open-vocabulary segmentation methods, we consider two open-sourced models that do not require mask supervision during training: MaskCLIP~\cite{zhou2022extract} and GroupViT~\cite{xu2022groupvit}.

For MaskCLIP, the target referring query and image are input to the model, with inference made similarly to our approach, \ie, obtaining prediction mask by thresholding the similarity map between all image patches and target query, using a threshold value of $0.5$.

For GroupViT, the target referring query and image are input to the model along with an additional dummy query (\eg, ``\texttt{A photo of a nothing}'').
This dummy query functions similarly to the background class in semantic segmentation methods. 
During inference, the model assigns image segments to the query with higher similarity, and segments assigned to the target query are considered the final prediction. 
We have also tried a similar inference protocol with our model for a fair comparison, but it failed to produce meaningful performance.

For the fine-tuning, we noticed that the loss does not decline with the original training configurations.
Therefore, we set the batch size and learning rate for both models to 32 and 1e-5, respectively, matching the hyperparameters used for our model.

\section{Decoder $f_\text{dec}$ for the reconstruction loss}
\label{sec:supp_decoder}
For the reconstruction decoder, we follow the model architecture of the spatial broadcast decoder~\cite{watters2019spatial} and its application to slot attention~\cite{seitzer2023bridging}.
Specifically, each entity slot is broadcasted to the $N$ sequence, which is the same length as the visual feature.
Next, the broadcasted slots are augmented with sinusoidal positional encoding.
These augmented slots are then fed into a four-layer multi-layer perceptron (MLP) with ReLU activation functions.
The output dimension of the MLP is $D+1$, with the additional dimension being used to compute weighting values via softmax.
Finally, we compute the reconstruction of the visual feature by taking the weighted sum of all slots in each sequence position, using the calculated weighting values.

\begin{table}[t!]
\setlength{\tabcolsep}{4pt}
\centering
\scalebox{0.8}{
\begin{tabular}{p{1.6cm}<{}|p{0.76cm}<{\centering}p{0.76cm}<{\centering}p{0.76cm}<{\centering}|p{0.76cm}<{\centering}p{0.76cm}<{\centering}p{0.76cm}<{\centering}|p{0.76cm}<{\centering}|p{0.76cm}<{\centering}}
\toprule
\multirow{2}{*}{Methods} 
& \multicolumn{3}{c|}{RefCOCO}
&\multicolumn{3}{c|}{RefCOCO+} 
&\multicolumn{1}{c|}{Gref}
&\multicolumn{1}{c}{PC} \\
                                                    & \textit{val} & \textit{testA} & \textit{testB}   & \textit{val} & \textit{testA} & \textit{testB}  & \textit{val} & \textit{val} \\ \midrule
\multirow{1}{*}{GroupViT}       & 10.82 & 11.11 & 11.29 & 11.14 & 10.78 & 11.84 & 12.77 & 9.41 \\  
\multirow{1}{*}{MaskCLIP}       & 19.45 & 18.69 & 21.37 & 19.97 & 18.93 & 21.48 & 21.11 & 23.80 \\ 
TSEG                            & 25.44 & - & - & 22.01 & - & - & 22.05 & 28.77\\ 
TSEG$\dagger$                   & 25.95 & - & - & 22.62 & - & - & 23.41 & 30.12\\\midrule
\multirow{1}{*}{Ours}           & 34.76 & 34.58 & 35.01 & 28.48 & 28.60 & 27.98 & 28.87  & 33.45 \\
\multirow{1}{*}{Ours$\dagger$}  & \textbf{35.75} & \textbf{35.52} & \textbf{36.03} & \textbf{29.30} & \textbf{29.41} & \textbf{28.67} & \textbf{30.02} & \textbf{35.67} \\
\bottomrule
\end{tabular}
}
\vspace{-1mm}
\caption{
Comparison with other methods, including the post-processing by DenseCRF.~\cite{Fullycrf}.
The results are reported in mIoU (\%). 
PC and $\dagger$ denote the PhraseCut dataset and the models post-processed by DenseCRF, respectively.
}
\vspace{-4mm}
\label{tab:quancrf}
\end{table}

\section{Additional qantitative results}
\label{sec:supp_more_quan}
Following the previous work that utilizes post-processing to enhance segmentation quality, we present the performance of our model with DenseCRF~\cite{Fullycrf} in~\Tbl{quancrf}.
The results demonstrate that employing DenseCRF yields performance improvements across all benchmarks.
Notably, the performance of our model without DenseCRF surpasses that of the previous method, TSEG~\cite{strudel2022weakly}, even when TSEG uses DenseCRF. 
In the main paper, we only report the performance of our model without DenseCRF, as its computation is time-consuming and the resulting benefits are relatively marginal.

\section{Additional qualitative results}
\label{sec:supp_more_qual}
In~\Fig{supp_qual1} and \Fig{supp_qual2}, qualitative results of our model on the RefCOCO \textit{val} set are presented.
The results demonstrate that our model, trained solely with image-text pair supervision, can successfully discover visual entities and integrate them into segmentation masks corresponding to free-form text queries. 
For instance, our model predicts accurate masks for referring expressions about non-human objects (row 2-3 in~Fig.~\ref{fig:supp_qual1},~\ref{fig:supp_qual2}), occluded objects (row 3 in~Fig.~\ref{fig:supp_qual1},~\ref{fig:supp_qual2}), and partially appeared objects (rows 5 in~Fig.~\ref{fig:supp_qual1},~\ref{fig:supp_qual2}).
Furthermore, the results of the top-3 discovered entities reveal that the entity discovery module effectively identifies visual entities, and the modality fusion module accurately infers their relevance to the text query.

In~\Fig{supp_qual3}, we present additional qualitative results of our framework, featuring three types of slots: entity (Ours), random, and query slots.
For the entity slots, those sampled from the same distribution are represented with identical boundary colors. 
The results indicate that using random slots leads to noisy entity discovery due to insufficient semantic specificity.
In contrast, utilizing query slots generates not adequately fine-grained entities, as they are bound to particular semantic categories.
Our proposed entity slot effectively addresses these shortcomings.
It produces accurate visual entities by maintaining an awareness of semantic properties and facilitating fine-grained entity discovery. 
Specifically, we can observe that the entity slots sampled from the same Gaussian distribution share the same semantic properties (\eg, head, chair, and color) while capturing individual entities without redundancy.

\clearpage

\begin{figure*}[t!]
    \centering
    \includegraphics[width=0.97\linewidth]{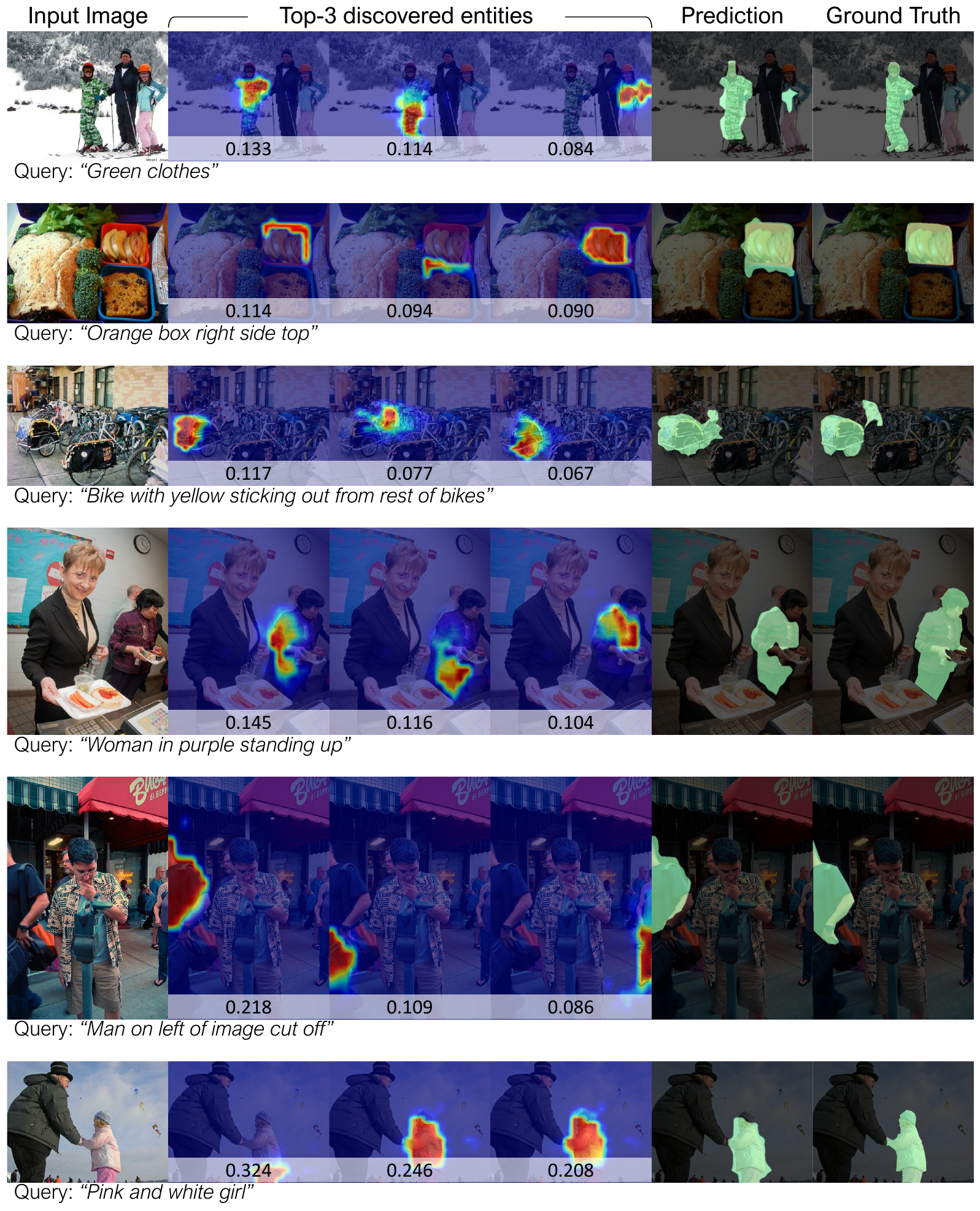}
    \caption{
    Qualitative results of our framework on RefCOCO \textit{val} set.
    We present the discovered entities from $A^\text{slot}$ and their relevance scores from $A^\text{fuse}$. Top-3 entities in terms of relevance to query expression are presented.
    } 
\label{fig:supp_qual1}
\end{figure*}

\begin{figure*}[t!]
    \centering
    \includegraphics[width=0.97\linewidth]{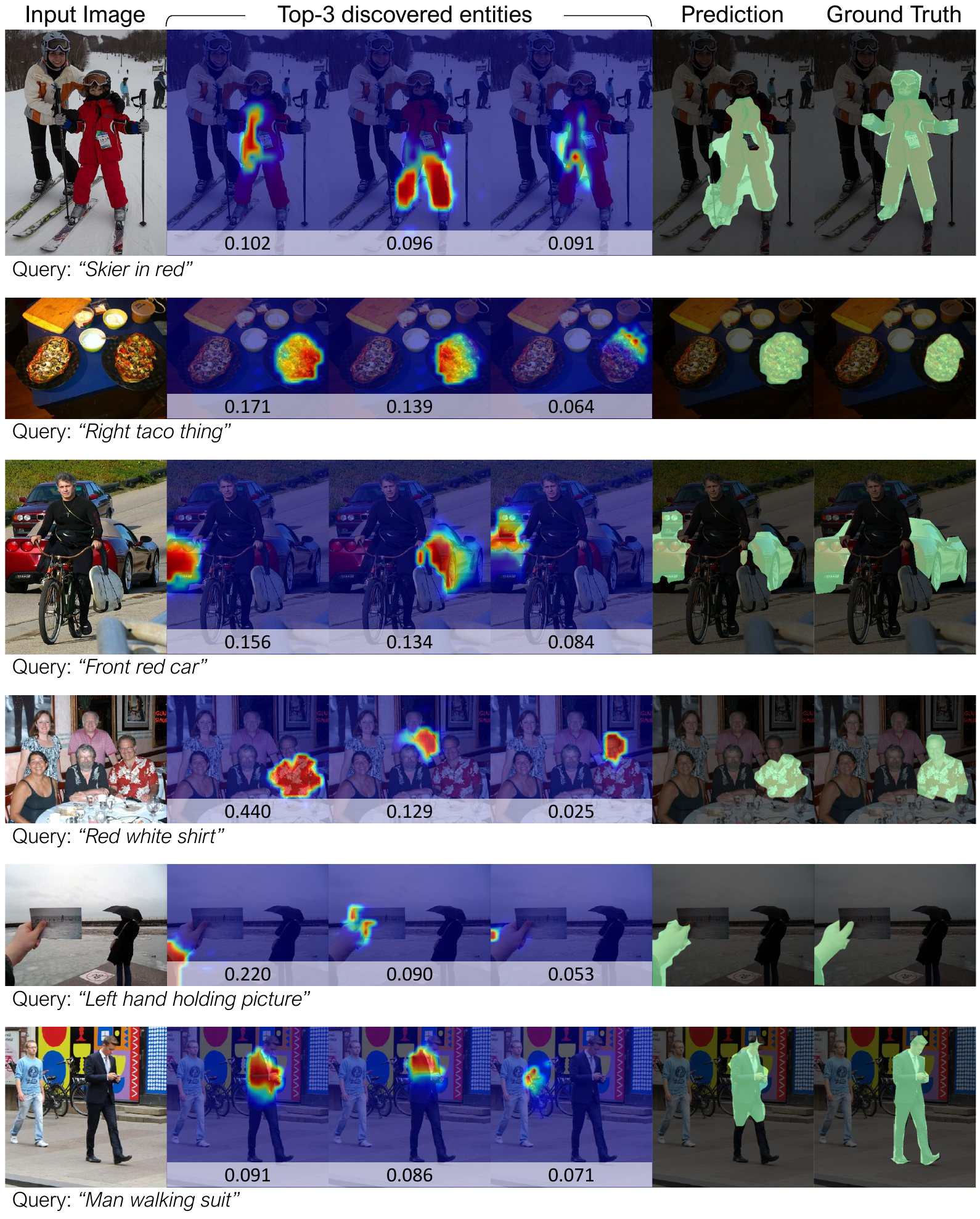}
    \caption{
    Qualitative results of our framework on RefCOCO \textit{val} set.
    We present the discovered entities from $A^\text{slot}$ and their relevance scores from $A^\text{fuse}$. Top-3 entities in terms of relevance to query expression are presented.
    } 
\label{fig:supp_qual2}
\end{figure*}

\begin{figure*}[t!]
    \centering
    \includegraphics[width=\linewidth]{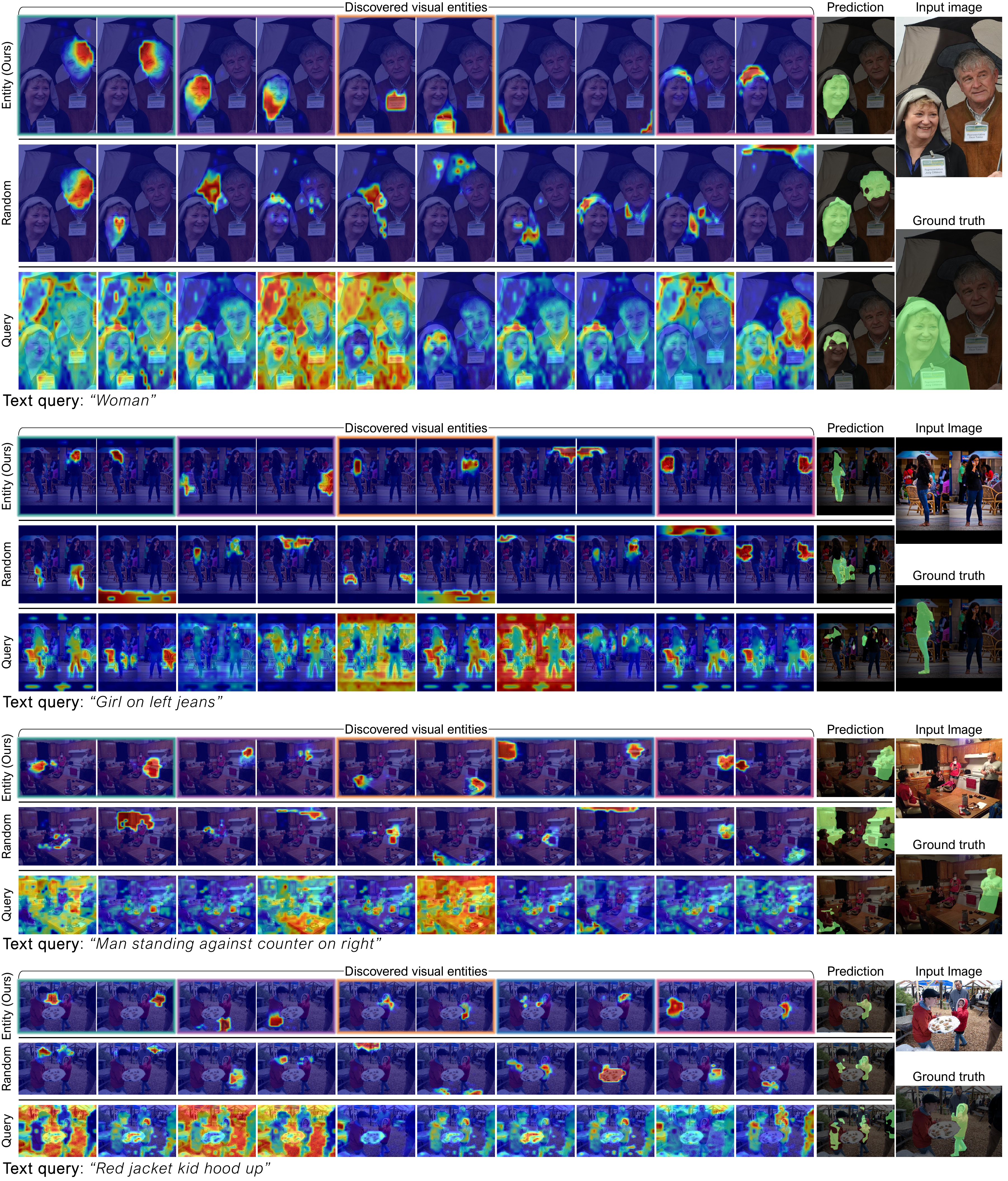}
    \caption{
    Qualitative results of our framework with entity slot (Ours), random slot, and query slot on RefCOCO \textit{val} set. 
    For each slot type, we present the 10 discovered entities from $A^\text{slot}$ and final predictions. 
    In the case of entity slots, the color of the boundaries indicates the Gaussian distribution that the slot sampled.
    } 
\label{fig:supp_qual3}
\end{figure*}
}

\end{document}